\pdfoutput=1 
\documentclass[10pt,twocolumn,letterpaper]{article}

\usepackage[pagenumbers]{cvpr} 

\usepackage[table]{xcolor}
\usepackage{multirow}
\usepackage{multicol}
\usepackage{hhline}
\usepackage{booktabs} 
\usepackage{arydshln}
\usepackage{array} 
\usepackage{stfloats}
\usepackage{amsmath}
\usepackage{pgfplots}

\usepackage{pgfplots}
\pgfplotsset{compat=1.17} 
\definecolor{modelcolor}{RGB}{173, 216, 230} 
\definecolor{lightyellow}{RGB}{255, 255, 224} 
\definecolor{lightpurple}{RGB}{230, 230, 250} 
\definecolor{checkmarkgreen}{RGB}{173, 255, 47} 

\definecolor{cvprblue}{rgb}{0.21,0.49,0.74}
\usepackage[pagebackref,breaklinks,colorlinks,citecolor=cvprblue]{hyperref}

\def\paperID{4512} 
\def\confName{CVPR}
\def\confYear{2024}

\title{EmoVIT: Revolutionizing Emotion Insights with Visual Instruction Tuning}
\author{Hongxia Xie\(^{1}\), Chu-Jun Peng\(^{2}\), Yu-Wen Tseng\(^{2}\),\\%
Hung-Jen Chen\(^{3}\), Chan-Feng Hsu\(^{3}\), Hong-Han Shuai\(^{3}\), Wen-Huang Cheng\(^{2}\)\\%
\(^{1}\)Jilin University\\%
\(^{2}\)National Taiwan University\\%
\(^{3}\)National Yang Ming Chiao Tung University}
\begin{document}
\maketitle

\begin{abstract}
    Visual Instruction Tuning represents a novel learning paradigm involving the fine-tuning of pre-trained language models using task-specific instructions. This paradigm shows promising zero-shot results in various natural language processing tasks but is still unexplored in vision emotion understanding.
    In this work, we focus on enhancing the model's proficiency in understanding and adhering to instructions related to emotional contexts. Initially, we identify key visual clues critical to visual emotion recognition. Subsequently, we introduce a novel GPT-assisted pipeline for generating emotion visual instruction data, effectively addressing the scarcity of annotated instruction data in this domain.
    Expanding on the groundwork established by InstructBLIP, our proposed EmoVIT architecture incorporates emotion-specific instruction data, leveraging the powerful capabilities of Large Language Models to enhance performance. Through extensive experiments, our model showcases its proficiency in emotion classification, adeptness in affective reasoning, and competence in comprehending humor.
    The comparative analysis provides a robust benchmark for Emotion Visual Instruction Tuning in the era of LLMs, providing valuable insights and opening avenues for future exploration in this domain. Our code is available at \url{https://github.com/aimmemotion/EmoVIT}.
\end{abstract}

\section{Introduction}
\label{sec:intro}
\begin{figure}[ht] 
\begin{flushleft} 
\begin{minipage}{0.5\textwidth} 
\centering
\includegraphics[scale=0.31]{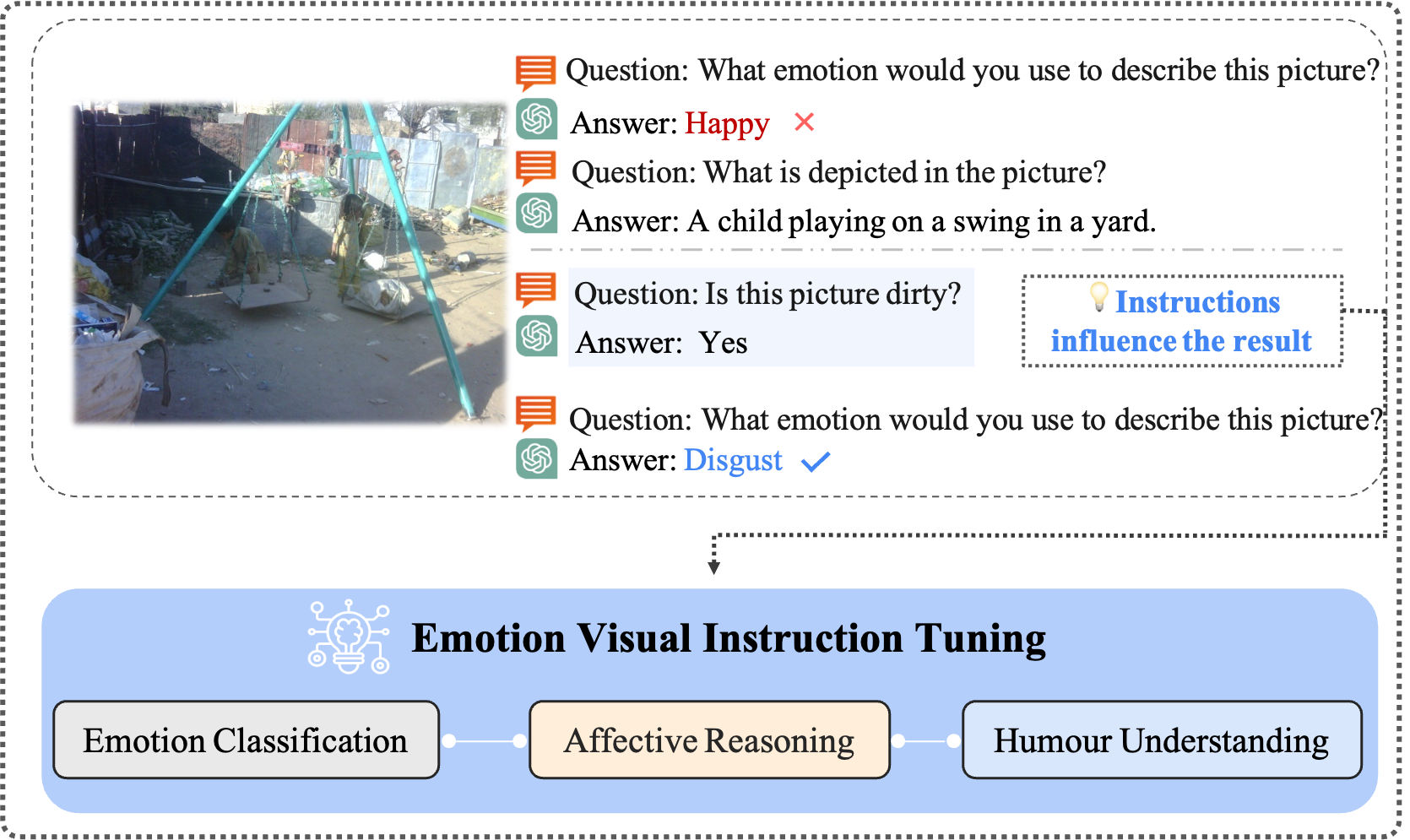}
\caption{Illustration of the importance of instruction-following ability in visual emotion understanding.}
\label{fig:intro}
\end{minipage}
\end{flushleft}
\vspace{-20pt} 
\end{figure}
Visual emotion recognition, a key area within artificial intelligence and computer vision, aims to predict human emotions based on visual cues such as facial expressions and body language. 
This technology is essential in bridging the gap between human affective states and machine understanding. 
Its diverse applications~\cite{xie2023most, yang2021stimuli_TIP, EmoSet, acii}, spanning from improving human-computer interaction to aiding in mental health assessment, underscore its significance. Accurate emotion recognition is vital for enhancing user experience and ensuring information security, as it helps prevent emotional manipulation and misinformation~\cite{euaiact}. 
Developing robust emotion recognition models is not only a technical challenge but also a step towards more empathetic and intuitive AI systems, paving the way for more efficient and natural human-computer interactions.



The AI community has recently shown a growing interest in developing foundational vision models, \textit{e.g.}, Flamingo~\cite{flamingo}, LLaVA~\cite{liu2023llava}, BLIP2~\cite{BLIP2}. 
These models excel in open-world visual understanding, tackling several vision tasks such as classification, detection, segmentation, and captioning. 
In contrast, current large-scale multimodal models are still in its infancy when it comes to emotion perception~\cite{lian2023explainable}. 
As illustrated in \cref{fig:intro}, when directly query the GPT-4~\cite{OpenAI2023GPT4} about the emotional category of an image, the model tends to provide incorrect responses. However, the model delivers accurate responses when provided with revised instructions.
To fully leverage the potential of existing vision-based large models, our approach is based on the concept of \textit{Instruction Tuning}. This effective strategy is aimed at teaching language models to follow natural language instructions, a technique proven to enhance their generalization performance across unseen tasks~\cite{FLAN_1st, liu2023llava,instructblip}.

In this work, we focus on \textit{developing the model's proficiency in understanding and following instructions related to emotional contexts}. This approach highlights the importance of fine-tuning the model's instruction-following capabilities, enabling it to interpret and respond to emotional content effectively. 
This is achieved by leveraging its pre-existing knowledge base, thereby eliminating the necessity for an emotion-specific architectural framework.

To address the notable challenges encountered in Instruction Tuning for visual emotion recognition, especially the lack of specific instruction data, we introduce a novel self-generation pipeline explicitly crafted for visual emotion recognition by using GPT-4~\cite{OpenAI2023GPT4}. 
This innovative pipeline excels in generating a diverse array of (image, instruction, output) instances, thereby notably enhancing the dataset with a more extensive and task-oriented variety of examples. This approach not only overcomes the challenge of limited data availability but also reduces the dependence on human labor. Therefore, it streamlines the process, enabling more efficient and effective emotion recognition.


Additionally, Instruction Tuning has been criticized for its emphasis on surface-level features like output patterns and styles, rather than achieving a profound comprehension and assimilation of tasks~\cite{InstructionTuning_survey}. 
To tackle this issue and enhance the diversity and creativity of instruction data, our dataset includes instructions that demand complex reasoning, going beyond basic question-and-answer formats.
This is further enriched by incorporating visual cues such as \textit{brightness}, \textit{colorfulness}, \textit{scene type}, \textit{object class}, \textit{facial expressions}, and\textit{ human actions}.
These aspects are pivotal in fostering a nuanced comprehension of visual emotions, thus allowing the model to generate more precise and contextually appropriate interpretations~\cite{EmoSet}.

After generating the emotion visual instruction data, we propose an Emotion Visual Instruction Tuning (EmoVIT) framework, leveraging the foundation of InstructBLIP~\cite{instructblip}. This framework incorporates an emotion-centric, instruction-aware module that proficiently guides Large Language Models (LLMs) in assimilating the nuances of emotion instructions. 
Our work signifies a paradigm shift, presenting a new era of instruction-based learning for visual emotion understanding that relies less on explicit training data.
Remarkably, as shown in \cref{fig:emoset}, our approach requires almost 50\% of the training data typically needed yet exceeds the performance of previous visual emotion recognition methods and popular Visual Instruction Tuning methods.




%
%

Our contributions can be summarized as follows:
\begin{itemize}
\item We explore the potential of the Visual Instruction Tuning paradigm for emotion comprehension and introduce the concept of Emotion Visual Instruction Tuning.
\item After thoroughly considering the unique characteristics of visual emotion recognition, we develop a novel GPT-assisted pipeline for generating emotion visual instruction data. This approach effectively bridges the gap in available annotated instruction data within this specific domain.
\item Building upon the foundation of InstructBLIP, our EmoVIT architecture integrates emotion domain-specific instruction data, harnessing the robust capabilities of LLMs to boost performance. The extensive experiments demonstrate our model's proficiency in emotion classification, affective reasoning, and comprehension of humour.

\end{itemize}

\definecolor{blue}{RGB}{135,206,235}
\definecolor{orange}{RGB}{255,165,0}
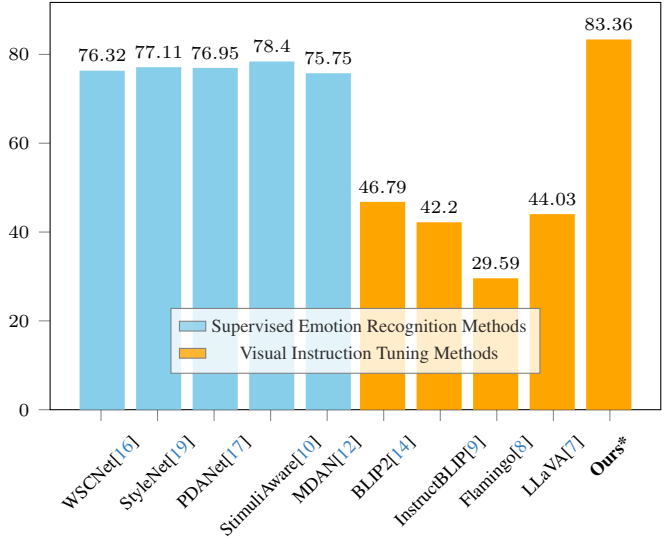
\begin{figure}
  \centering
  \begin{tikzpicture}
    \begin{axis}[
      ybar, 
      bar width=0.6cm, 
      width=9.7cm,
      height=7cm,
      xtick={1,2,3,4,5,6,7,8,9,10}, 
      xticklabels={WSCNet\cite{wscnet}, StyleNet\cite{StyleNet}, PDANet\cite{PDANet},StimuliAware\cite{yang2021stimuli_TIP}, MDAN\cite{MDAN}, BLIP2\cite{BLIP2}, InstructBLIP\cite{instructblip}, Flamingo\cite{flamingo}, LLaVA\cite{liu2023llava}, \textbf{Ours*}}, 
      ymin=0, 
      xticklabel style={rotate=45, font=\scriptsize},
      nodes near coords, 
      nodes near coords align={vertical}, 
      nodes near coords style={font=\scriptsize}, 
      y tick label style={font=\scriptsize}, 
      ytick align=outside, ytick pos=left,
      xtick pos=bottom,
      legend style={at={(0.5, 0.25)}, anchor=north, fill opacity=0.8, font=\scriptsize, draw=gray}, 
      legend entries={Supervised Emotion Recognition Methods, Visual Instruction Tuning Methods}, 
      legend image code/.code={
        \draw[#1] (0cm,-0.05cm) rectangle (0.35cm,0.08cm);
      },
    ]
  
    \addplot [draw=none, fill=blue, bar shift=0.015cm] coordinates {(1, 76.32) (2, 77.11) (3, 76.95) (4, 78.4) (5, 75.75) };
    \addplot [draw=none, fill=orange, bar shift=-0.015cm] coordinates {(6, 46.79)(7, 42.2) (8, 29.59) (9, 44.03) (10, 83.36)};
  
    \end{axis}
  \end{tikzpicture}
  \caption{Performance comparison on EmoSet test set~\cite{EmoSet} (Accuracy \%).}
  \label{fig:emoset}
\end{figure}

\section{Related Work}
\label{sec:relatedwork}

\subsection{Visual Emotion Recognition}
A key challenge in visual emotion recognition is bridging the gap between an image's visual cues and the emotions it portrays~\cite{MDAN,ECCV18_BIAS,fi}. While traditional efforts, \textit{e.g.}, Xu \textit{et al.}'s multi-level dependent attention network~\cite{MDAN}, focus on visual models for emotional feature learning, recent advancements like EmoSet~\cite{EmoSet} offer rich emotion-laden datasets with 3.3 million images. The rise of multimodal models, such as the GPT series~\cite{OpenAI2023GPT4}, has further propelled Vision-Language Recognition. However, fully leveraging these models in emotion recognition is an area ripe for exploration. Our work leads the way in utilizing large-scale models for Emotion Visual Instruction Tuning.

\subsection{Visual Instruction Tuning}

Current Large Language Models (LLMs) have extensive knowledge bases, but their effectiveness depends on accurately interpreting human instructions due to a mismatch between training goals and user expectations. LLMs are trained to minimize prediction errors, whereas users expect helpful and safe instruction-following. Instruction Tuning addresses this by teaching models to follow natural language instructions, enhancing generalization to new tasks. FLAN~\cite{FLAN_1st} demonstrated that training a large model on instruction-based datasets improves zero-shot performance. This approach has extended to vision-language tasks, with BLIP2~\cite{BLIP2} and LLaVA~\cite{liu2023llava} adapting instruction-tuned LLMs for visual inputs. InstructBLIP~\cite{instructblip} introduces instruction-aware visual feature extraction and the Q-Former, enabling more flexible, instruction-driven feature extraction.

As a novel area, visual emotion instruction tuning lacks benchmarks or guidelines for creating emotion instruction data. Our work pioneers the use of large-scale models to develop an emotion instruction data pipeline, overcoming the limitations of manual annotation.

\section{Method}
\label{sec:method}

\subsection{Preliminary of Visual Instruction Tuning}

\begin{figure*}[bpt]
\begin{center} 
\begin{minipage}{1\textwidth} 
\centering
\includegraphics[scale=0.8]{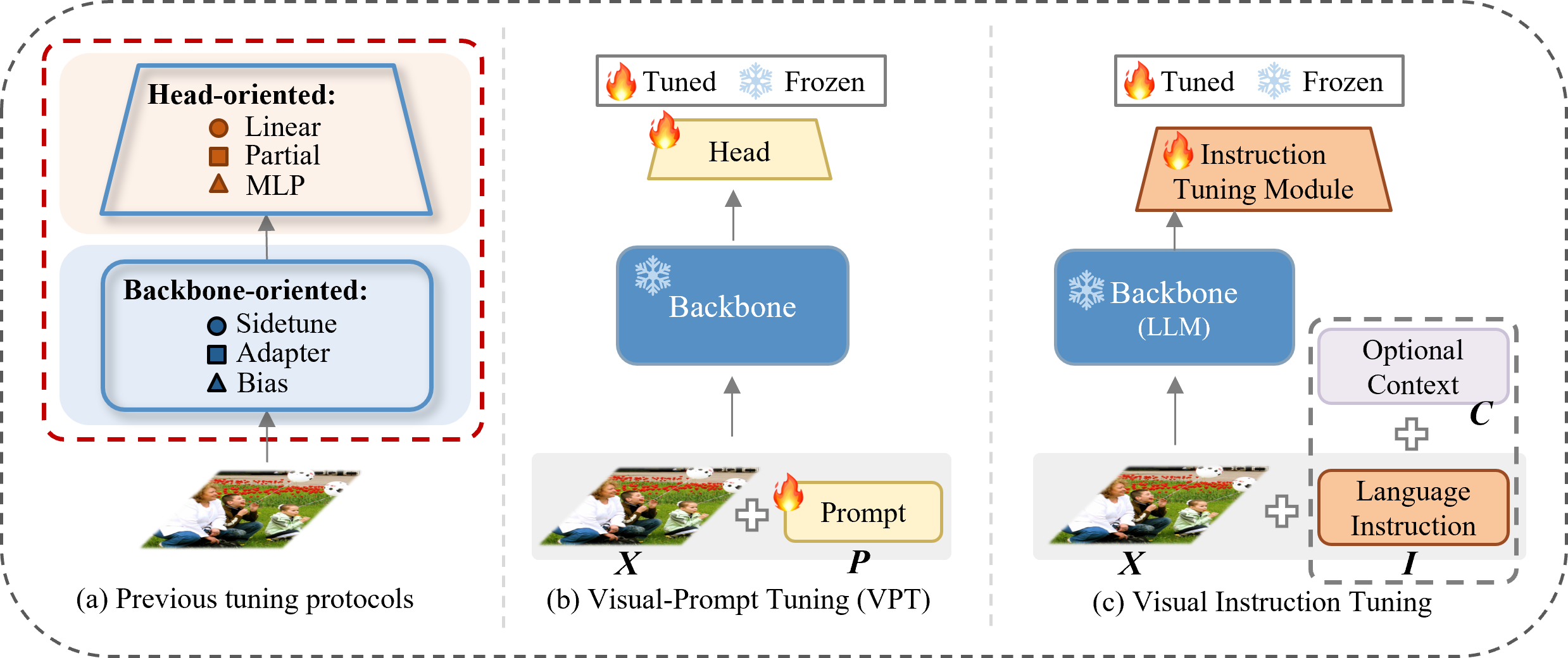}
\caption{The comparison of different visual tuning paradigms.}
\label{fig:viusal_tuning_comparison}
\end{minipage}
\end{center}
\vspace{-20pt} 
\end{figure*}

In the deep learning era, \textbf{visual tuning} has experienced significant paradigm shifts, as depicted in \cref{fig:viusal_tuning_comparison}. 

In \cref{fig:viusal_tuning_comparison}\textcolor{red}{(a)}, conventional tuning methodologies encompass \textit{Full fine-tuning, Head-oriented, and Backbone-oriented} techniques, capitalizing on large-scale pre-trained models. Predominantly, thoroughly fine-tuning these models for specific tasks, conducted end-to-end, is recognized as a highly effective strategy. However, this method requires maintaining separate copies of the backbone parameters for each distinct task, posing challenges in storage and deployment.

Alternatively, Visual Prompt Tuning (VPT)~\cite{visual_prompt_tuning}, presents an efficient substitute for full fine-tuning within large-scale vision Transformer models. It achieves this by employing a minimal fraction of trainable parameters in the input space while maintaining a frozen backbone model. 
The objective function for Visual Prompt Tuning is given by:
\begin{equation}
   \min_{\theta_{\text{P}}} \mathcal{L}(f(X, P; \theta_{\text{P}}), Y)
\end{equation}
where $\min_{\theta_{\text{P}}}$ is the minimization over the prompt parameters $P$, $\mathcal{L}$ is the loss function, $f$ represents the model function with input image $X$, prompt parameters $P$, and learnable model parameters $\theta_{\text{P}}$ as input, and $Y$ is the target output.


Visual Prompt Tuning focuses on optimizing LLMs using a small set of parameters, whereas Visual Instruction Tuning (VIT) aims to improve the model's comprehension of instructions, thereby addressing the model's shortcomings in specific domains.
This type of method aims to enhance the model's proficiency in following instructions, leveraging the capabilities of the latest foundation models, \textit{e.g.}, Llama~\cite{Llama}, and BLIP2~\cite{BLIP2}. 
Instructions serve as guiding constraints, shaping the model's outputs to conform to specific response characteristics and domain-relevant knowledge.
This approach enables human monitoring of the model's behavior, thereby assuring alignment with the desired outcomes. Moreover, Instruction Tuning is computationally efficient, allowing LLMs to swiftly adapt to particular domains without extensive retraining or architectural alterations. 


The objective function for Visual Instruction Tuning is given by:
\begin{equation}
     \min_{\theta_{\text{tunable}}} \mathcal{L}(g(X, I, C; \theta_{\text{tunable}}), Y)
\end{equation}
where $\min_{\theta_{\text{tunable}}}$ denotes the minimization over the tunable parameters $\theta_{\text{tunable}}$ in the Instruction Tuning Module, $\mathcal{L}$ is the loss function, $g$ is the model function with instruction $I$, image $X$, other contexts $C$, and tunable parameters $\theta_{\text{tunable}}$, and $Y$ denotes the target output. The optional context $C$ is not just raw data; it encompasses descriptive or directive information guiding the model on how to process input or which task to execute, \textit{e.g.,} image caption. It's integral to the model's understanding and execution of tasks based on specific instructions or guidelines.

\begin{figure*}[t] 
\begin{flushright} 
\begin{minipage}{1\textwidth} 
\centering
\includegraphics[scale=0.65]
{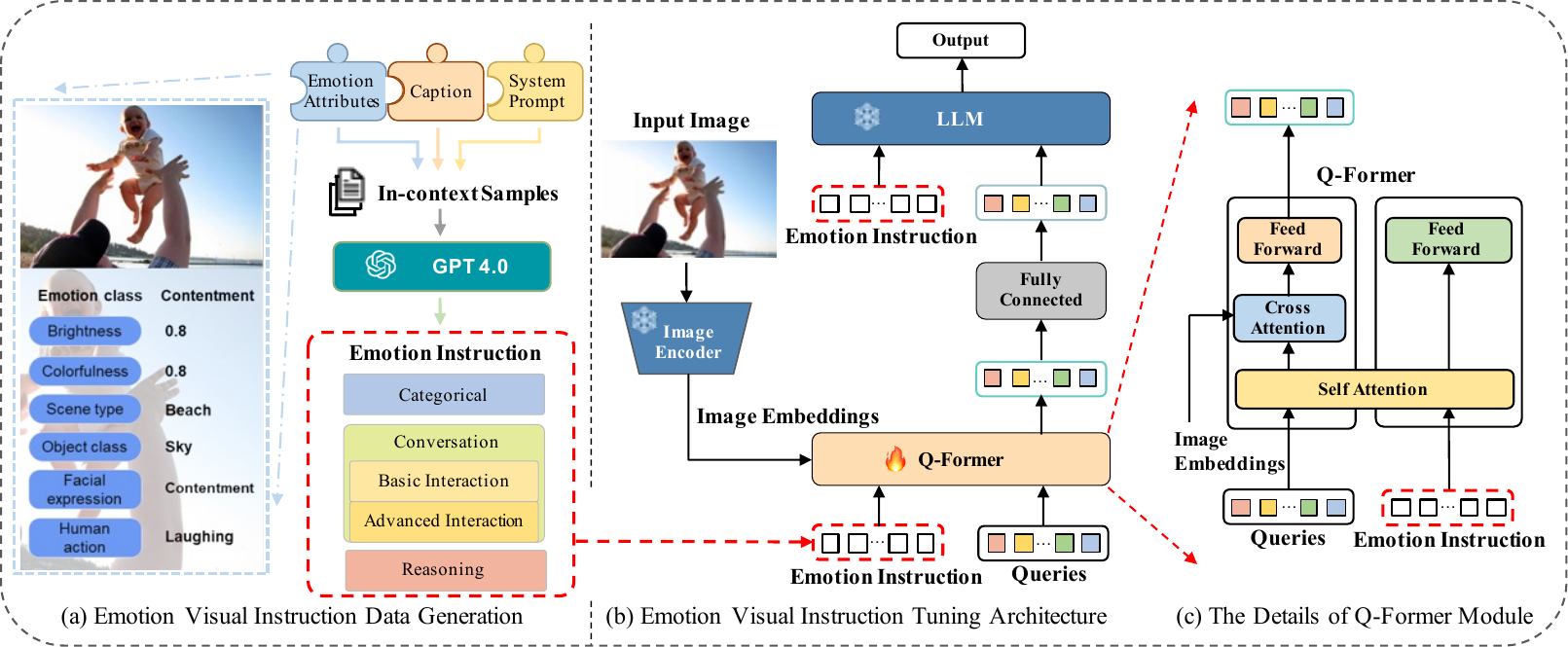}
\caption{The overall architecture of our proposed method. The \textit{Emotion Instruction} data generated by (a) will be used for \textit{Emotion Visual Instruction Tuning} in (b). During \textit{Emotion Visual Instruction Tuning}, given an input image, the frozen Image Encoder initiates the process by extracting visual features. Emotion Instruction generated by (a) are subsequently interacting with \textit{Queries} embedding through the learnable Q-Former. This interaction is key to drawing out image features that are relevant to the task at hand. As a result, the frozen LLM receives visual information conducive to instruction following.}
\label{fig:overall_architecture}
\end{minipage}
\end{flushright}
\vspace{-20pt} 
\end{figure*}

\subsection{GPT-assisted Emotion Visual Instruction Data Generation}
\label{sec:gpt-assisted data generation}

Previous methodologies commonly employed a consistent template-based set of instructions for every image within a dataset across various specific tasks~\cite{instructblip}.
For instance, a standard instruction such as ``\textit{Briefly describe the content of the image}'' was employed uniformly across all images for Image Captioning. 
In this way, the model may not be able to adequately capture the unique characteristics of each image. 
Moreover, this \textit{one-size-fits-all} approach often leads to suboptimal performance in emotion recognition tasks that require nuanced perception and differentiation of ambiguous emotion classes.


Since the topic of Emotion Visual Instruction Tuning is still in its infancy, no benchmarks or guidelines have been proposed so far for constructing emotion instruction data. 
Based on the recent successes of machine-generated instructions demonstrated in LLaVA~\cite{liu2023llava}, our work pioneers the use of existing LLMs to create a pipeline for self-generating emotion instructions.
Different from previous template-based and \textit{one-size-fits-all} instruction data, we propose an instance-wise and LLM-assisted visual emotion instruction data pipeline.
This methodology transcends the constraints of manual annotation by employing GPT-4~\cite{OpenAI2023GPT4} to generate instance-wise, tailored instruction data that dynamically corresponds to visual content. 

Prior to the development of instructional data for the visual emotion recognition task, it is imperative to confront a fundamental academic problem: \textit{What types of visual clues are pivotal in identifying emotions?} 
This necessitates a careful consideration of the unique characteristics inherent to the task, along with a comprehensive understanding of the potential visual cues associated with human emotions.
In this work, we propose a novel visual instruction data mechanism to remove the inherent subjectivity and ambiguity in emotional interpretation. 
Specifically, we integrate a broad spectrum of emotion attributes across multiple levels: \textit{low-level} attributes (\textit{e.g.}, brightness, colorfulness), \textit{mid-level} attributes (\textit{e.g.}, scene type and object class), and \textit{high-level} attributes (\textit{e.g.}, facial expressions and human actions), building upon insights from previous work~\cite{EmoSet}. 
This comprehensive strategy not only aligns with the intricate nature of emotions but also significantly enhances the model's capability to interpret and understand visual emotional cues more accurately and holistically.




The overall pipeline of our proposed emotion visual instruction data is shown in \cref{fig:overall_architecture} \textcolor{red}{(a)}.
For an image \(X_{\text{img}}\), three types of image-related contexts are essential for GPT-4 to generate emotion instruction data: \( (i) \) a caption \(X_c\), \( (ii) \) an emotion attribute list \(X_{\text{attr}}\), which includes \textit{emotion class, brightness, colorfulness, scene type, object class, facial expression}, and \textit{human action}, and \( (iii) \) the system prompt, designed to enable GPT-4 to comprehend the specific task requirement\footnote{A detailed description of the system prompt is provided in the supplementary materials.}.

\begin{figure}[t] 
\begin{flushleft} 
\begin{minipage}{0.5\textwidth} 
\centering
\includegraphics[scale=0.5]{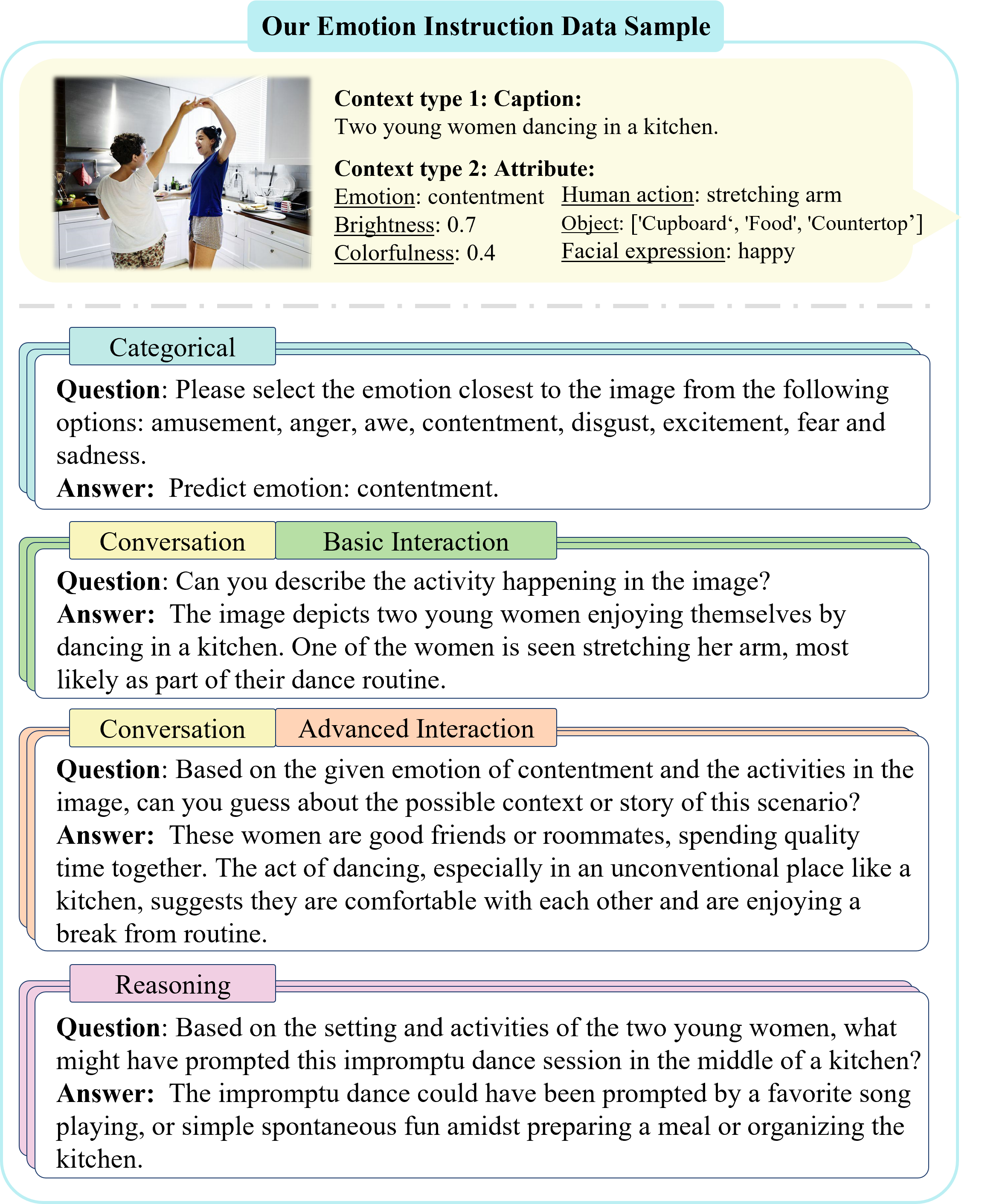}
\caption{The sample of our generated visual emotion instruction data.}
\label{fig:instruction_sample}
\end{minipage}
\end{flushleft}
\vspace{-20pt} 
\end{figure}

We first manually design a few examples which are used as seed examples for in-context learning to query GPT-4. This operation leverages the model's ability to extrapolate from given examples, enhancing its understanding and response accuracy based on the principles of few-shot learning~\cite{liu2023llava}.
Our generated emotion instruction data includes three types: \textit{Categorical}, \textit{Conversation}, and \textit{Reasoning}. 
Building upon previous research~\cite{liu2023llava}, our generated instruction data adheres to the dialogue format, exemplified in \cref{fig:instruction_sample}. 

Our strategy for generating emotion instruction data adopts a progressive approach \textit{from simple to complex}. Initially, for the \textit{Categorical} data, we transform the associated emotion class of the image into a structured format. This process serves as the foundational component of our emotion instruction data.


For the \textit{Conversation} data, our framework is designed to create dialogues in which the GPT assistant interacts with an inquirer, focusing on the emotion attributes of the image. In this setup, the assistant's responses are tailored to interpret and describe the image as though it were within its own visual field, thereby providing insights from an observational viewpoint. The scope of questions posed is comprehensive, encompassing the types of objects depicted, their actions, and the dynamics of their interrelationships.
The dialogues we generate fall into two categories:  \( (i) \) \textit{Basic Interaction}, focusing on the provided emotion attribute list with simple, direct characteristics, and \( (ii) \) \textit{Advanced Interaction}, which builds on the first type to reach greater conversational complexity and sophistication.

For the \textit{Reasoning} data, our approach extends beyond mere visual content, prompting the model to generate in-depth reasoning questions.
To enhance the dialogue's credibility and structure, detailed examples are incorporated alongside logical reasoning steps, ensuring that the discourse convincingly captures the intricacies of the visual content.



\subsection{Emotion Visual Instruction Tuning }


After acquiring the emotion visual instruction data as detailed in \cref{sec:gpt-assisted data generation}, our goal is to employ this data in enhancing the existing Visual Instruction Tuning model. This enhancement aims to align the LLMs' existing knowledge with the emotion understanding domain.


As shown in \cref{fig:overall_architecture} \textcolor{red}{b}, we have developed an Emotion Visual Instruction Tuning (EmoVIT) architecture based on InstructBLIP~\cite{instructblip}.
This architecture specifically leverages its Instruction-aware Q-Former Module, as depicted in \cref{fig:overall_architecture} \textcolor{red}{c}, for emotion-centric instructional tasks.

Specifically, the Instruction-aware Q-Former Module takes in the \textit{emotion instruction tokens}, \textit{queries}, and \textit{image embeddings} as input. 
The image embeddings are extracted by a frozen image encoder. The learnable queries are initially produced by the pre-trained Q-Former of InstructBLIP. 
During training, the Instruction-aware module enhances task-specific feature extraction. It does this by integrating emotion instruction and query embeddings within self-attention layers, aligning visual information with the LLM's instruction-following requirements.
Our approach adopts cross-entropy loss, tailoring it to the intricacies of visual emotion recognition tasks, thus ensuring precise and contextually relevant model training outcomes.



We note that the data generated by our approach is not confined to a single model but can also be applied to other Visual Instruction Tuning models, such as LLaVA~\cite{Llama}. Notably, when LLaVA is fine-tuned with our data, it exhibits a significant enhancement in emotion recognition capabilities, as detailed in \cref{sec:held-out}. In this way, we demonstrate not only the effectiveness but also the transferability of our generated data, showing its broad applicability and impact.

\section{Experimental Results}
\label{sec:experiments}
\subsection{Implemental Details}



Our implementation is based on the LAVIS library~\cite{li2022lavis}. 
Our EmoVIT starts with a pre-trained InstructBLIP baseline and proceeds to fine-tune exclusively the Q-Former module, whilst keeping both the image encoder and the language model frozen.
The parameters for our training adhere to the default settings established by InstructBLIP.

\noindent\textbf{Datasets.}
We evaluate our framework on ten benchmark datasets annotated under different scenarios and class number, namely EmoSet~\cite{EmoSet}, WEBEmo~\cite{ECCV18_BIAS}, Emotion6~\cite{emotion6}, the Flickr and Instagram (FI)~\cite{fi}, Artphoto~\cite{artphoto_abstract}, IAPS~\cite{IAPSA}, Abstract~\cite{artphoto_abstract}, EmotionROI~\cite{emotionroi}, UnbiasedEmo~\cite{ECCV18_BIAS}, and OxfordTVG-HIC~\cite{OxfordTVG_HIC}.

\noindent{\textbf{Held-in Pretraining.}} Following previous work~\cite{instructblip}, we divide our dataset into two categories: held-in for pretraining and held-out for evaluation~\footnote{Unlike the setup in InstructBLIP, our dataset exclusively comprises emotion-related content. Consequently, our held-out evaluation does not constitute a strict zero-shot evaluation in the conventional sense.}.
Considering the EmoSet dataset's comprehensive inclusion of emotion attributes for each image, it has been chosen as the primary resource for our held-in pretraining phase. Simultaneously, for a broader assessment, we perform held-out evaluations using the test sets from various other datasets.

For the generation of emotion visual instruction data, we initially employ the BLIP2 model for image captioning, followed by leveraging the GPT-4 API to generate emotion instruction data.
In total, our collection comprises \textit{Categorical}, \textit{Conversation}, and \textit{Reasoning} instruction data, derived from 51,200 unique images. This represents less than 50\% of the entire EmoSet.

\subsection{Held-out Evaluation}
\label{sec:held-out}
As shown in \cref{tab:main}, our proposed methodology exhibits a marked superiority in performance relative to the 
burgeoning Visual Instruction Tuning Methods. 
Since they have been pre-trained on dozens of large-scale datasets, it is evident that our generated emotion visual instruction data is particularly effective for emotional understanding
Our results signify a paradigm shift, heralding a new era of model training that relies less on explicit supervision and more on the robustness of emotion instruction-driven learning.

\begin{table*}
    \centering
    \begin{tabular}{lccccccccccc}
        \toprule
       \textbf{ Method} & \textbf{WebEmo} & \textbf{FI} & \textbf{Emotion6} & \textbf{Abstract} & \textbf{ArtPhoto} & \textbf{IAPSa} & \textbf{EmotionROI}& \textbf{EmoSet}\\
       Number of Classes &25&8&6&8&8&8&6&8&\\
        \midrule 
        Flanmingo~\cite{flamingo}  &9.36
&14.91&21.67&3.57
 & 17.5
 & 10.13 & 21.72
 & 29.59 
 \\
        LLaVA~\cite{liu2023llava}  &12.55
&56.04&49.44
&19.54
  & 36.25 & 42.43 &46.46 & 44.03
\\
       BLIP2~\cite{BLIP2}  &20.10
&57.72&50.00& 28.57 & 36.25 &39.24 & 50.51 & 46.79
 \\
        InstructBLIP~\cite{instructblip} &12.80
&37.97&46.11&21.42& 26.25 & 34.18& 46.13 & 42.20
 \\
 
        \midrule
        \cellcolor{gray!25} \textbf{Ours*}  &\cellcolor{gray!25} \textbf{21.12}&\cellcolor{gray!25} \textbf{68.09}&\cellcolor{gray!25} \textbf{57.81}& \cellcolor{gray!25} \textbf{32.34}  & \cellcolor{gray!25} \textbf{44.90}  & \cellcolor{gray!25} \textbf{44.13} & \cellcolor{gray!25} \textbf{53.87}  & \cellcolor{gray!25} \textbf{83.36}  \\
        \bottomrule
    \end{tabular}
     \caption{Held-out performance comparison on visual emotion datasets (\%).}
     \label{tab:main}
\end{table*}

\begin{figure}[t] 
\begin{center} 
\begin{minipage}{0.5\textwidth} 
\centering
\includegraphics[scale=0.55]{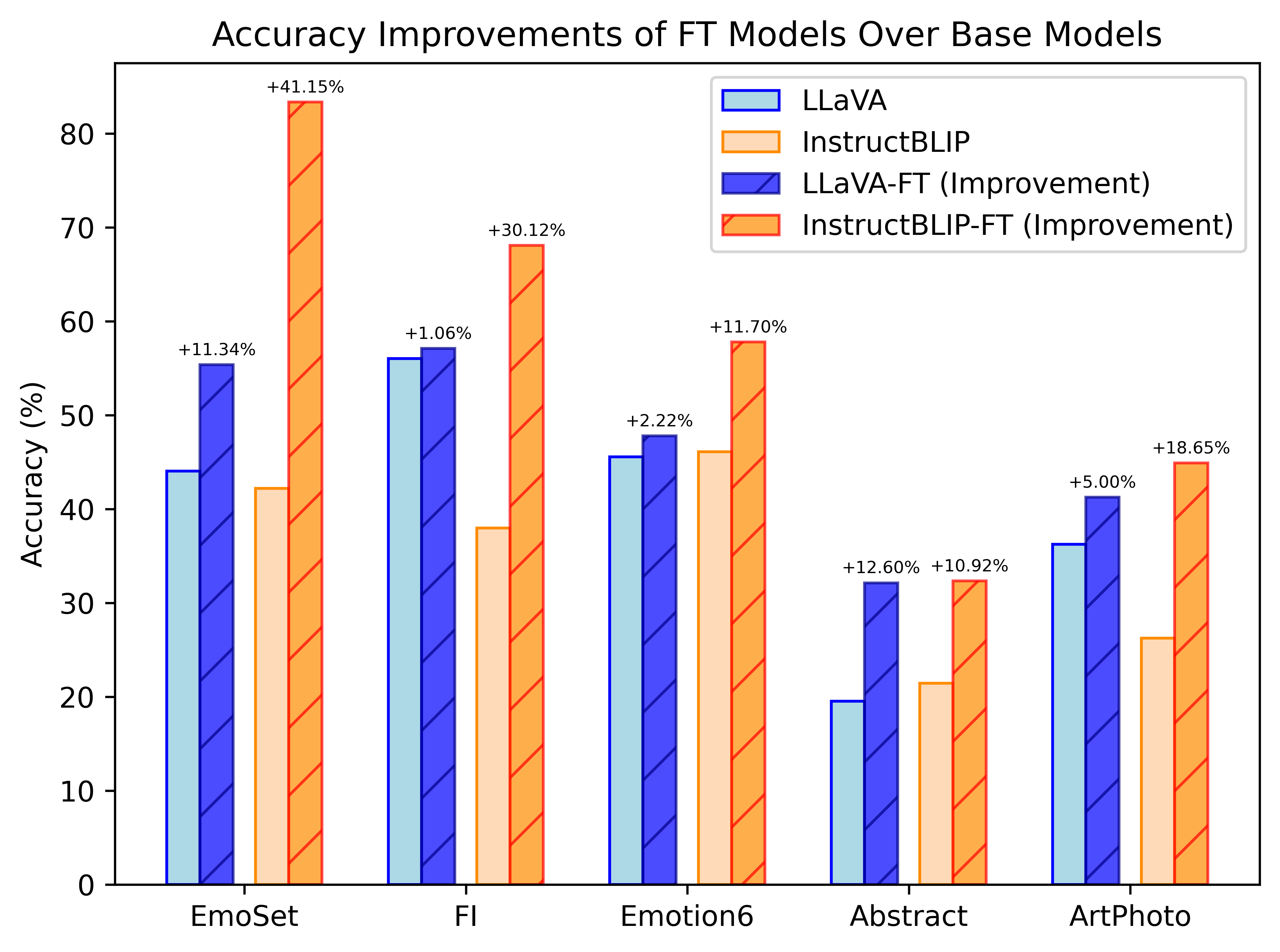}
\caption{The improvement of our proposed emotion visual instruction tuning data tuning on LLaVA~\cite{liu2023llava} and InstructBLIP~\cite{instructblip}.}
\label{fig:tuning_improvement}
\end{minipage}
\end{center}
\vspace{-20pt} 
\end{figure}

\textbf{The Effectiveness of Our Proposed Emotion Visual Instruction Data.}
As the first to introduce the concept of emotion visual instruction data, our study seeks to evaluate the generalizability of this newly generated instruction data. Our goal is to test its efficacy not only with InstructBLIP but also across other Visual Instruction Tuning model, to understand its broader applicability.
As depicted in \cref{fig:tuning_improvement}, we employ two Visual Instruction Tuning models, LLaVA and InstructBLIP, which were fine-tuned on our specially generated emotion visual instruction data. Subsequent testing across five distinct datasets reveals notable improvements in both models, substantiating the efficacy of our generated data. Notably, InstructBLIP demonstrated a more substantial overall enhancement compared to LLaVA. This can be attributed to InstructBLIP's specialized Instruction-aware Q-Former Module, which adeptly extracts the salient features of our emotion instructions and synergizes them effectively with the corresponding images, thereby yielding improved performance.

\subsection{Effectiveness of Different Instruction Data}

\subsubsection{Ablation Study of Different Instruction Data}

The ablation study outlined in \cref{tab:ablation_diff_instruction} provides a comprehensive analysis of the impact that different instructional data types have on model performance, specifically concerning accuracy metrics on the EmoSet test set. Initially, the model, referred to as InstructBLIP~\cite{instructblip}, operates without the integration of the three types of instructional data and attains a baseline accuracy of 42.20\%. This foundational performance is significantly enhanced with the inclusion of \textit{Categorical} data, which alone contributes to a substantial increase in accuracy. The introduction of \textit{Conversation} data further amplifies this effect, underscoring the value of conversational context in improving the model's predictive capabilities.
The addition of \textit{Reasoning} data notably boosts performance, achieving a peak accuracy of 83.36\%. This indicates that the model significantly benefits from the nuanced cues in reasoning, aiding in understanding complex emotional instructions. The gradual improvements with each data type support the idea that a diverse approach to instructional data markedly enhances model comprehension and performance.


\begin{table}[ht]
\centering
\small
\begin{tabular}{cccc}
\hline
 \textbf{Categorical } & \textbf{Conversation } & \textbf{Reasoning } & \textbf{Accuracy (\%)} \\ \hline
 - & - & - & 42.20\\ \hdashline
\checkmark & - & - & 80.90 (\textcolor{blue}{+38.70})\\
\checkmark & \checkmark & - & 81.95 (\textcolor{blue}{+39.75}) \\
\checkmark & \checkmark & \checkmark & 83.36 (\textcolor{blue}{+41.16})  \\ \hline
\end{tabular}
\caption{Ablation study of three types of instruction data. Accuracy (\%) on EmoSet test set. }
\label{tab:ablation_diff_instruction}
\vspace{-5mm}
\end{table}

\subsubsection{Instruction Sensitivity}

This work is dedicated to the creation of a varied corpus of visual emotion instruction data, alongside the development of a robust instruction-based model. Our objective is for the model to demonstrate stability, producing consistent results in the face of minor variations in instruction phrasing, provided the core objective of the task persists unchanged.
To this end, we employ the \textit{Sensitivity} evaluation metric, as introduced by~\cite{xu2022multiinstruct}, to assess the model's fidelity in generating uniform outcomes irrespective of instructional nuances.

We employ two semantically similar instructions as input prompts for the model, testing their impact on the Sensitivity score across three visual emotion datasets for different Visual Instruction Tuning models. The first instruction is:
``From the given options: \texttt{cls\_1}, \texttt{cls\_2}, \texttt{cls\_3}, etc., identify the emotion that most accurately reflects the image. Ensure your selection is confined to the listed options. Respond in the format: Predicted emotion:''
The second one states:
``Please choose the emotion that best corresponds to the image from the following options: \texttt{cls\_1}, \texttt{cls\_2}, \texttt{cls\_3}, etc. (Do not provide answers beyond the provided candidates.) Please reply in the following format: Predict emotion:''

\begin{figure}[t] 
\begin{center} 
\begin{minipage}{0.5\textwidth} 
\centering
\includegraphics[scale=0.6]{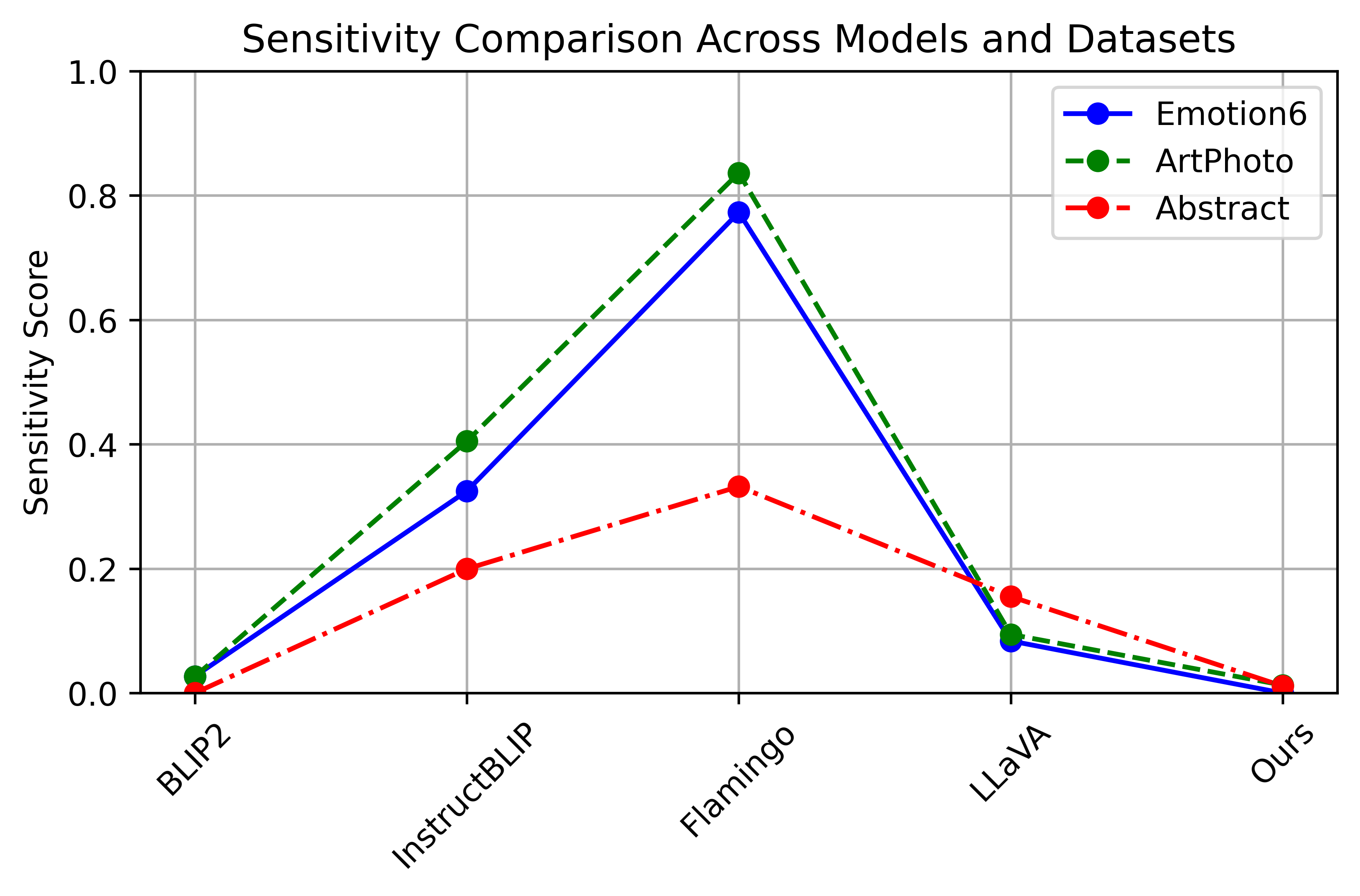}
\caption{The sensitivity score comparison (the lower the better).}
\label{fig:sensitivity}
\end{minipage}
\end{center}
\vspace{-5mm}
\end{figure}

As illustrated in \cref{fig:sensitivity}, our approach, along with BLIP2, exhibited exceptionally low Sensitivity values, demonstrating robustness in understanding the instructions. Conversely, Flamingo and InstructBLIP displayed a higher degree of sensitivity, indicating a relative susceptibility to variations in instruction wording.


\subsection{Robustness}
Given that current emotion recognition datasets often exhibit category imbalances and labeling biases, our aim is to evaluate the generalization ability of various learning strategies more impartially. Hence, we selected the UnBiasedEmo test set~\cite{ECCV18_BIAS}, which is uniquely suited for recognizing intricate emotions, such as those associated with identical objects or scenes, \textit{e.g.}, landscapes, crowds, families, babies, and animals, where the emotional undertones can be particularly subtle and complex. 

As depicted in \cref{tab:UnbiasedEmo}, our proposed methodology demonstrates superior performance when benchmarked against conventional supervised emotion recognition techniques, thereby underscoring the efficacy of our approach in more accurately discerning complex emotional contexts.

\begin{table}[ht]
\centering
\begin{tabular}{lc}
\toprule
\textbf{Method} & \textbf{Accuracy (\%)} \\
\midrule
Direct Learning~\cite{ECCV18_BIAS}  & 71.64 \\
Self-Directed Learning~\cite{ECCV18_BIAS}  & 72.45 \\
Joint Learning ~\cite{ECCV18_BIAS} & 71.64 \\
Curriculum Learning~\cite{ECCV18_BIAS} & 74.27 \\
\textbf{Ours*} & \textbf{74.72}\\
\bottomrule
\end{tabular}
\caption{Performance comparison on UnbiasedEmo dataset.}
\label{tab:UnbiasedEmo}
\end{table}

\subsubsection{Affective Reasoning}
In the domain of visual emotion recognition, where ambiguity and subjectivity are pervasive, the advent of an interpretable model is of considerable value. Such a model elucidates its cognitive processes, enhancing its trustworthiness and practicality in scenarios requiring a delicate grasp of emotional subtleties.

Leveraging Visual Instruction Tuning, our model transcends mere categorization of emotions; it articulates the underlying rationale for its classifications. The executing commands for identifying emotions and elucidating the decision basis is illustrated below:

\begin{verbatim}
      Predicted emotion: [emotion]. 
      Reason: [explanation].
\end{verbatim}

Our model delineates the visual features influencing its determinations, thereby addressing the complexities inherent in discerning and explaining emotion-related nuances.

\begin{figure}[t] 
\begin{center} 
\begin{minipage}{0.5\textwidth} 
\centering
\includegraphics[scale=0.7]{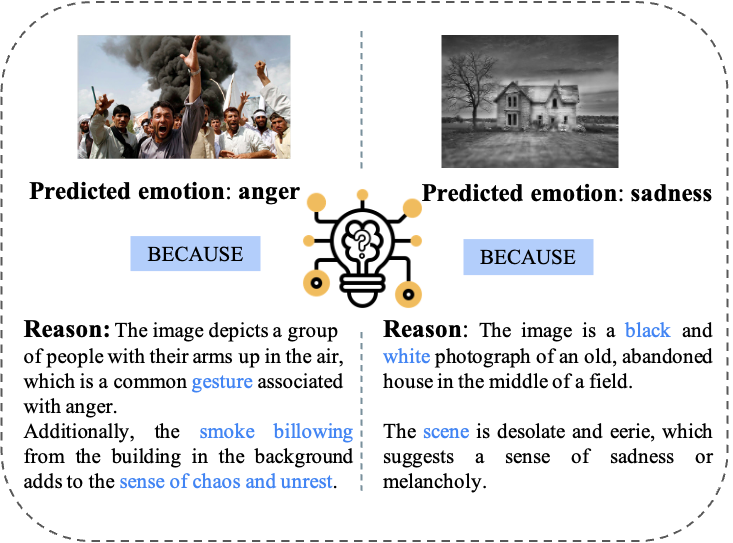}
\caption{The sample of our generated explanation.}
\label{fig:reasoning_sample}
\end{minipage}
\end{center}
\vspace{-10mm}
\end{figure}

The explanations provide us with visual clues contained within the images, as exemplified in~\cref{fig:reasoning_sample}. It provides interpretable visual indicators that inform the model's outputs, as demonstrated in our example, by disambiguating the often abstract emotional categories.

\subsection{Scaling Law}
\noindent\textbf{Pretraining data.} 
As demonstrated in \cref{tab:ablation_pretraining_data}, there is a clear correlation between the size of the pre-training dataset and improved performance. Consequently, we anticipate that an increase in training data in the future could enhance the effectiveness of Emotion Visual Instruction Tuning.
\begin{table}[ht]
\centering
\small
\begin{tabular}{cccc}
\hline
 \textbf{5\%
 } & \textbf{10\%
 } & \textbf{30\%
 } & \textbf{50\%} \\ \hline
 79.00 & 81.00 & 79.34 & 83.36\\ \hline
\end{tabular}
\caption{Ablation study of different portion of pre-training data. Accuracy (\%) on EmoSet test set. }
\label{tab:ablation_pretraining_data}
\end{table}




\subsection{Humour Caption Generation}
\begin{figure}[t] 
\begin{center} 
\begin{minipage}{0.5\textwidth} 
\centering
\includegraphics[scale=0.3]{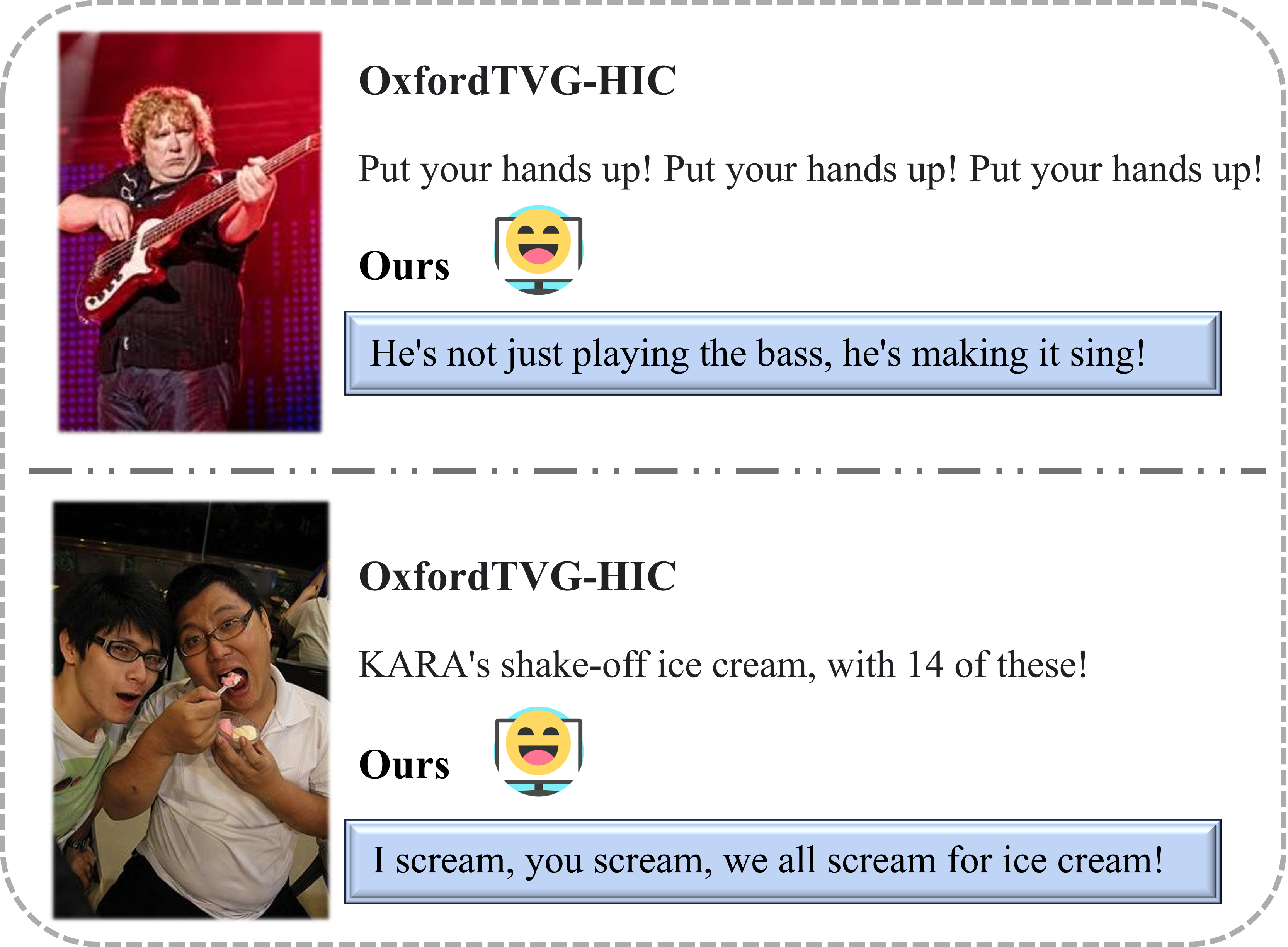}
\caption{The sample of our generated humour caption \textit{vs} human writing humour caption from OxfordTVG-HIC.}
\label{fig:humour_sample}
\end{minipage}
\end{center}
\vspace{-10mm}
\end{figure}

The comprehension of humor is intricately linked to the understanding of emotions.
Leveraging our generative language model, we conduct a caption generation task without modifying the model's architecture, specifically testing the model's proficiency in generating humorous captions. For this purpose, we select 50 images from the OxfordTVG-HIC dataset~\cite{OxfordTVG_HIC} and generate corresponding captions using our model.
Subsequently, the captions produced by our model are compared with manually annotated captions from the dataset in a user study. Thirty participants were asked to vote on which captions were more humorous. Our model-generated captions receive 60\% of the votes, demonstrating its effective humor generation capabilities. One sample is visualized in \cref{fig:humour_sample}.

\section{Conclusion}
\label{sec:conclusion}

In our study, drawing upon the distinctive visual cues key to visual emotion recognition, we present a GPT-assisted pipeline specifically designed for generating emotion visual instruction data. The developed EmoVIT model incorporates emotion-specific instructions, leveraging LLMs for enhanced performance. Our comprehensive experiments validate its effectiveness in emotion classification, affective reasoning, and humor understanding. This comparative analysis sets a benchmark for Emotion Visual Instruction Tuning with LLMs, providing valuable insights and directions for future research in this field.

\clearpage
\setcounter{page}{1}
\maketitlesupplementary

\setcounter{figure}{9}
\setcounter{table}{4}



\section{More Emotion Visual Instruction Data Sample}

\begin{figure}[t] 
\begin{flushleft} 
\begin{minipage}{0.5\textwidth} 
\centering
\includegraphics[scale=0.5]{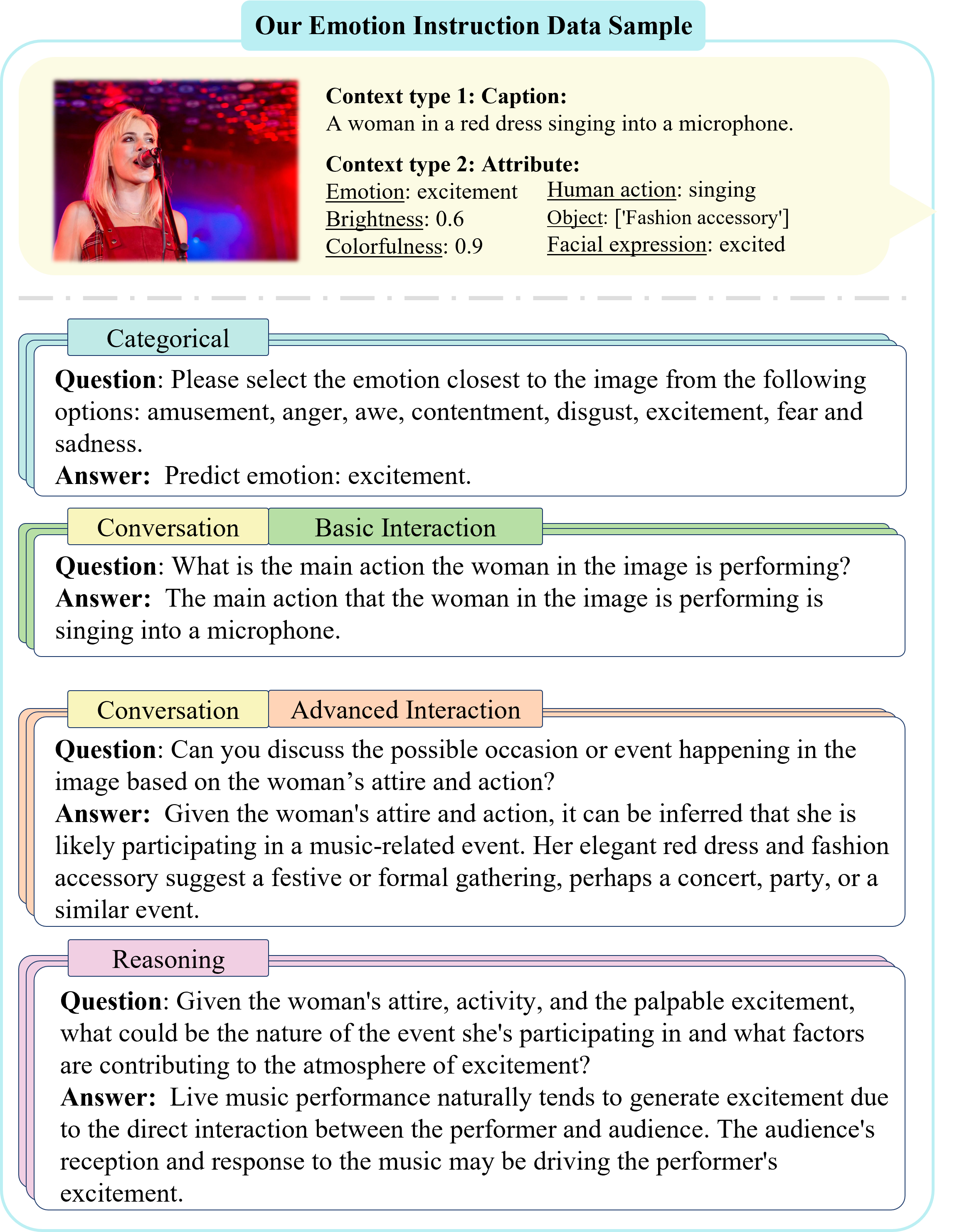}
\caption{The sample of our generated visual emotion instruction data.}
\label{fig:instruction_sample_1}
\end{minipage}
\end{flushleft}
\vspace{-20pt} 
\end{figure}

\begin{figure}[t] 
\begin{flushleft} 
\begin{minipage}{0.5\textwidth} 
\centering
\includegraphics[scale=0.5]{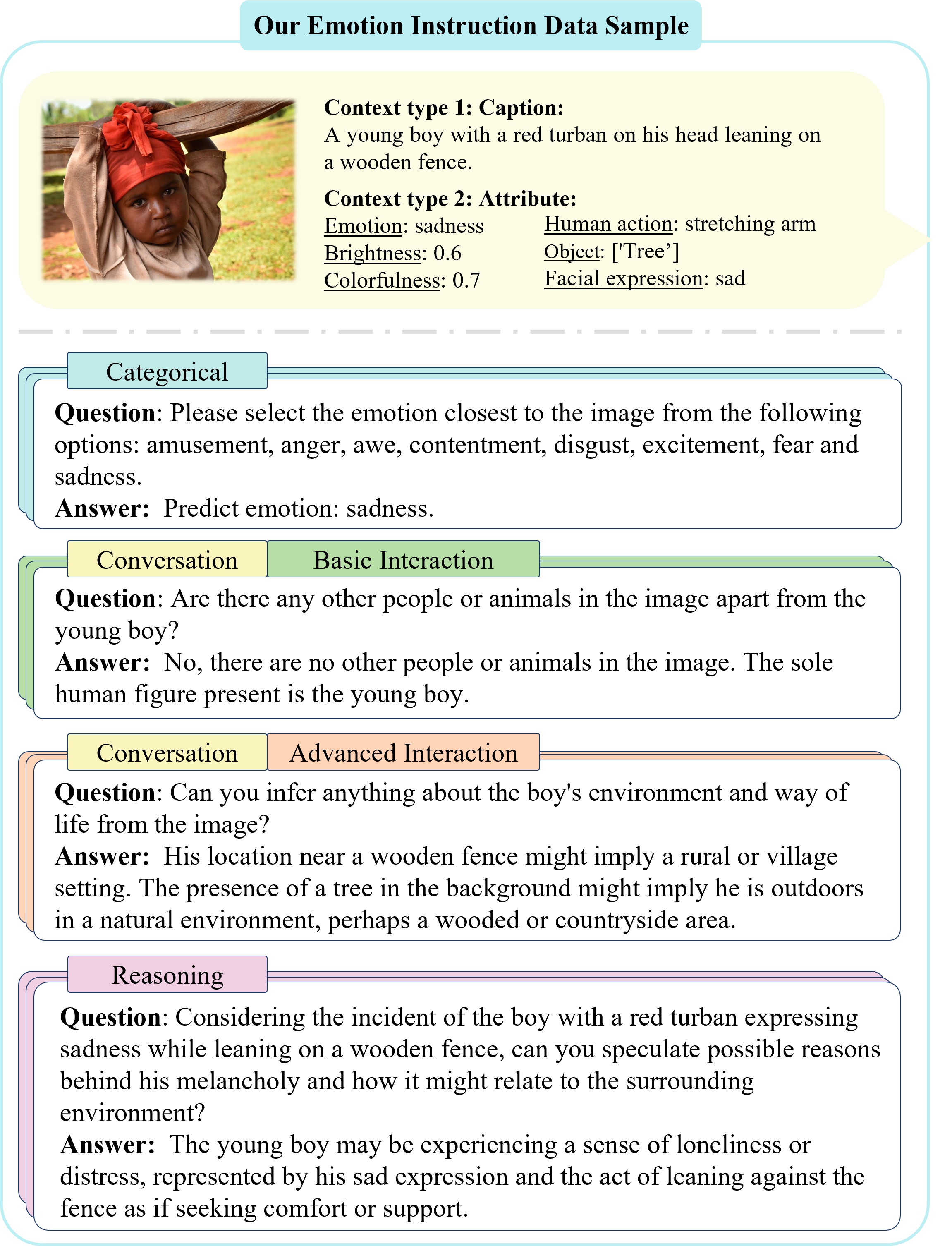}
\caption{The sample of our generated visual emotion instruction data.}
\label{fig:instruction_sample_2}
\end{minipage}
\end{flushleft}
\vspace{-20pt} 
\end{figure}

Additional samples from our Emotion Visual Instruction Data collection are presented in Figures \ref{fig:instruction_sample_1} and \ref{fig:instruction_sample_2}. Upon acceptance, the complete dataset will be made available on our project webpage.

\section{Implemental Details}
\subsection{Our Experiment Settings}
\textit{Held-out} vs \textit{supervised learning}.
We adopt the terminology\textit{ held-in} and \textit{held-out} as defined in the work of InstructBLIP~\cite{instructblip}. For the held-in, we utilize the training subset of the EmoSet dataset for Emotion Visual Instruction Tuning, with its corresponding test subset serving the purpose of held-in evaluation. The outcomes of this evaluation are depicted in Fig. \textcolor{red}{1} of the main manuscript.

In our held-out evaluation, we focus on determining how instruction tuning bolsters the model’s ability to transfer learning to new and unseen data. It’s crucial to highlight that our methodology sets a distinct path from InstructBLIP's framework. Our dataset is specifically curated with emotion-centric content, presenting unique categories such as \textit{cheerfulness} and \textit{enthrallment} found in WEBEmo, which are not typically included in other datasets. Conversely, common emotional categories like \textit{anger} and \textit{fear} are shared with other collections, such as FI and Emotion6. This distinctive mix in our dataset implies that our held-out evaluation operates on a \textit{cross-domain} level, examining the model’s ability to interpret and adapt to diverse emotional contexts not strictly confined to zero-shot scenarios.

\subsection{System Prompt}
The system prompt inputted into ChatGPT for the purpose of gathering instruction-based data is presented below.
\begin{quotation}
\textit{You are an AI visual assistant, and you are seeing a single image. What you see are provided with one caption and some emotion related attributes, describing the same image you are looking at. Answer all questions as you are seeing the image.
The range of brightness is from 0 (darkest) to 1 (brightest), and the range of colorfulness is from 0 (black-and-white) to 1 (the most colorful).}

\textit{Design two questions for a conversation between you and a person asking about this photo. The answers should be in a tone that a visual AI assistant is seeing the image and answering the question.
Ask diverse questions and give corresponding answers.}

\textit{Include questions asking about the visual content of the image, including the object types, object actions, relationship among objects, etc. Only include questions that have definite answers:
(1) one can see the content in the image that the question asks about and can answer confidently;
(2) one can determine confidently from the image that it is not in the image.
Do not ask any question that cannot be answered confidently.
Please answer with the format
Question:
Answer:}

\textit{Also include one complex question that is relevant to the content in the image, for example, asking about background knowledge of the objects in the image, asking to discuss about events happening in the image, etc. Again, do not ask about uncertain details.
Provide detailed answers when answering complex questions. For example, give detailed examples or reasoning steps to make the content more convincing and well-organized.  You can include multiple paragraphs if necessary.}

\end{quotation}
\subsection{Details of the Q-Former}
Similar to the approach in InstructBLIP, Q-Former is a lightweight transformer architecture that utilizes a collection of trainable query vectors to distill visual features from a static image encoder. The Q-Former acts as the trainable module to bridge the
gap between a frozen image encoder and a frozen LLM. Its role is to curate and present the most pertinent visual information, thereby enabling the LLM to generate the targeted textual output efficiently. Following the default setting, in our experimental setup, we employ 32 distinct queries, each with a dimensionality of 768. 
\subsection{Sensitivity Formula}
As mentioned in Sec.\textcolor{red}{4.3.2} in the main paper, we employ the \textit{Sensitivity} evaluation metric, as introduced by~\cite{xu2022multiinstruct}, to assess the model's fidelity in generating uniform outcomes irrespective of instructional nuances.
Specifically, for each task \( t \in T \), given its associated instances with task instructions: \( D^t = \{ (I_j^t, x_j^t, y_j^t) \in T \times X^t \times Y^t \}_{j=1}^N \), sensitivity is defined as:
\begin{equation}
\mathbf{E}_{t \in T} \left[ \frac{
\sigma_{i \in I^t} \left[ \mathbb{E}_{(x,y) \in D^t} \left[ \mathcal{L}(f_{\theta}(i, x), y) \right] \right]
}{
\mu_{i \in I^t} \left[ \mathbb{E}_{(x,y) \in D^t} \left[ \mathcal{L}(f_{\theta}(i, x), y) \right] \right]
} \right]
\end{equation}
where \(\mathcal{L}\) denotes the evaluation metric, \textit{i.e.}, emotion classification accuracy, \(f_{\theta}(\cdot)\) represents the Visual Instruction Tunign model. The standard deviation and mean of the model's performance across all instructions are denoted by \(\sigma_{i \in I^t}[\cdot]\) and \(\mu_{i \in I^t}[\cdot]\), respectively.

\section{Ablation Study of LLM Model Size}
In our attempts with the EmoVIT architecture's LLM, we explored the use of models of varying sizes (as shown in \cref{tab:ablation_model_size}). The results indicated that the smaller model, Vicuna7B, outperformed its larger counterparts. This may be attributed to the limited training data available for our task, which potentially underutilizes the capabilities of larger models. Consequently, we anticipate that an increase in training data in the future could enhance the effectiveness of Emotion Visual Instruction Tuning.

\begin{table}[ht]
\centering
\small
\begin{tabular}{ccc}
\hline
 \textbf{Vicuna-7B
 } & \textbf{Vicuna-13B
 } & \textbf{FlanT5XL
 }  \\ \hline
 83.36 & 82.21 & 80.98\\ \hline
\end{tabular}
\caption{Ablation study of different LLM model size. Accuracy (\%) on EmoSet test set. }
\label{tab:ablation_model_size}
\end{table}

\section{GPT-4 \textit{vs} GPT-4 Turbo}

We conducted a comparative analysis of conversational datasets derived from GPT-4 (the model name is \textit{gpt-4} in the API) against the recently released GPT-4 Turbo (the model name is \textit{gpt-4-1106-preview} in the API). The comparative metrics yielded negligible differences between the two models (83.36\% \textit{vs} 82.96\% on EmoSet test set).

\section{Adding In-context Samples in Held-out Evaluation}
Recent LLMs are capable of in-context learning when provided with a limited number of examples in a few-shot manner. In this work, we have also embarked on such an exploration. 
For instance, \cref{tab:emotion_descriptions} presents the in-context samples utilized within the EmotionROI dataset. During our held-out evaluation, we incorporated three in-context samples for each category, consisting of a caption paired with its corresponding emotion class.
Nevertheless, in our experimental observations, we did not witness any enhancement in performance attributable to furnishing the LLM with these in-context examples.
Consequently, our finalized methodology did not incorporate in-context samples during the held-out evaluation phase.
\begin{table*}[ht]
\centering
\begin{tabular}{p{0.9\linewidth}l}
\toprule
\textbf{Description} & \textbf{Emotion} \\
\midrule
Unleashed Fury: A portrait of raw, unfiltered anger etched on the subject's face. & Anger \\
Volcanic Eruption in Human Form: A Portrait of Unrestrained Fury. & Anger \\
An explosive portrait of raw fury, where every clenched jaw and furrowed brow tells a tale of unchecked anger. & Anger \\
Face contorted in a grimace of pure disgust, as if they just tasted a year-old lemon. & Disgust \\
Caught in the throes of revulsion, a face grimaces as if it just tasted the world's sourest lemon. & Disgust \\
Picture Perfect: A Masterclass in the Art of Disgust Expression & Disgust \\
A chilling moment of pure terror, etched in every detail. & Fear \\
A chilling moment of pure terror etched on the face, a stark embodiment of fear. & Fear \\
someone with a wide smile, a group & Joy \\
Overflowing with joy, like a puppy at a park! & Joy \\
A poignant portrait of sorrow, where teardrops are the silent language of grief. & Sadness \\
An evocative portrayal of sorrow, with shadows seemingly swallowing the light, reflecting the heavy weight of sadness. & Sadness \\
An abstract portrayal of solitude, where the vivid hues of melancholy paint a poignant picture of sadness. & Sadness \\
Caught in a moment of pure astonishment, eyes wide and mouth agape. & Surprise \\
Caught in the headlights of astonishment: a jaw-dropping moment of surprise! & Surprise \\
Caught in the Act! A person's wide-eyed gasp of sheer surprise. & Surprise \\
\bottomrule
\end{tabular}
\caption{Illustrative Examples of Emotion Descriptors in Visual Data}
\label{tab:emotion_descriptions}
\end{table*}

\section{Limitation and future work}
Due to the reliance on the GPT-API and cost considerations, our held-in pretraining phase utilized less than 50\% of the EmoSet dataset. Despite outperforming other methods, we recognize the potential for significant improvements in future work by expanding the data scale. We anticipate that advancements in visual emotion understanding will parallel increases in both data and model scale.
{
    \small
     \clearpage
     \clearpage
    \bibliographystyle{ieeenat_fullname}
    \bibliography{main}
}


\end{document}